\pdfoutput=1
\documentclass[11pt]{article}

\usepackage[final]{acl}
\usepackage{times}
\usepackage{latexsym}
\usepackage[T1]{fontenc}
\usepackage[utf8]{inputenc}
\usepackage{microtype}
\usepackage{inconsolata}
\usepackage{graphicx}
\usepackage{subcaption}
\usepackage{color}
\definecolor{dkgreen}{rgb}{0,0.6,0}
\definecolor{gray}{rgb}{0.5,0.5,0.5}
\definecolor{mauve}{rgb}{0.58,0,0.82}
\usepackage{listings}
\usepackage{amsmath}
\usepackage{paralist}
\usepackage{url}
\usepackage{hyperref}

\title{taz2024full: Analysing German Newspapers for Gender Bias and Discrimination across Decades}

\author{
 \textbf{Stefanie Urchs\textsuperscript{1,2}},
 \textbf{Veronika Thurner\textsuperscript{1}},
 \textbf{Matthias Aßenmacher\textsuperscript{2,3}},
 \textbf{Christian Heumann\textsuperscript{2}},
\\
 \textbf{Stephanie Thiemichen\textsuperscript{1}},
 \\
\\
 \textsuperscript{1}Faculty for Computer Science and Mathematics,\\ Hochschule München University of Applied Sciences,
 \textsuperscript{2}Department of Statistics, LMU Munich,\\
 \textsuperscript{3}Munich Center for Machine Learning (MCML), LMU Munich,
\\
 \small{
   \textbf{Correspondence:} \href{mailto:stefanie.urchs@hm.edu}{stefanie.urchs@hm.edu}
 }
}

\begin{document}
\maketitle
\begin{abstract}
Open-access corpora are essential for advancing natural language processing (NLP) and computational social science (CSS). However, large-scale resources for German remain limited, restricting research on linguistic trends and societal issues such as gender bias. We present \texttt{taz2024full}, the largest publicly available corpus of German newspaper articles to date, comprising over 1.8 million texts from taz, spanning 1980 to 2024.

As a demonstration of the corpus’s utility for bias and discrimination research, we analyse gender representation across four decades of reporting. We find a consistent overrepresentation of men, but also a gradual shift toward more balanced coverage in recent years. Using a scalable, structured analysis pipeline, we provide a foundation for studying actor mentions, sentiment, and linguistic framing in German journalistic texts.

The corpus supports a wide range of applications, from diachronic language analysis to critical media studies, and is freely available to foster inclusive and reproducible research in German-language NLP.
\end{abstract}

\section{Introduction}
Understanding how language reflects and shapes society requires access to large, well-structured text collections. While English dominates the landscape of publicly available corpora, resources for other languages—such as German—remain comparatively scarce. This limits the scope of language-specific research on topics like linguistic change, media discourse, and social bias. In particular, studying gender discrimination in news media calls for datasets that span long time periods and include sufficient volume and variation to analyse these phenomena \textit{at scale}.

Given that the term "\textit{gender}" is not always used uniformly, we define its usage in the scope of this paper: We refer to "\textit{gender}" as a social construct, which is non-binary, flexible, shaped by individuals and by those perceiving them, rather than a biological given \cite{doi:10.1177/0891243287001002002,10.1145/3531146.3534627}. However, due to methodological and linguistic constraints, we are limited to working within a binary gender spectrum in this work. We use pronouns to assume the gender of actors. As we are working on German data, we are limited to the commonly used pronouns in this language. Since German lacks widely used non-binary pronouns, non-binary genders could not be included in our analysis. We further employ the definition of "\textit{biases}" as “all notions and beliefs a person has towards another person or group of persons” \cite{mateo2020more}. These biases can manifest in (written) discrimination, wherein a person or group is intentionally mistreated based on specific characteristics \cite{Reisigl2017}.

\textbf{Contribution:} In this work, we present a twofold contribution to the research area of gender bias and gender discrimination in written language and potentially also beyond:
\begin{enumerate}
    \item We introduce the corpus ``\texttt{taz2024full}'', a comprehensive dataset that can serve as a valuable resource for analysing German newspaper articles published between 1980 and 2024, a period spanning more than four decades. Beyond our use case (see below), this allows analysing various other phenomena in the German language over time.
    \item Furthermore, we demonstrate the potential of the corpus for bias and discrimination research by analysing how references to different genders have developed over more than 40 years of newspaper reporting, examining both their frequency and how they are discussed.
\end{enumerate}

\section{Related Work}
Most publicly available news corpora focus on English newspapers, providing this type of diverse natural language processing (NLP) research resource only for a single language. The "\textit{Chronicling America}" dataset, provided by the Library of Congress, offers access to historical newspapers and digitised pages (spanning from 1690 to the present day)~\cite{Library_of_Congress}. The "\textit{BBC News Summary Dataset}" includes 2,225 documents from 2004 to 2005, covering five topical areas~\cite{10.1007/978-981-16-9012-9_21}. The "News Category Dataset"\footnote{\url{https://huggingface.co/datasets/heegyu/news-category-dataset}} contains around 210,000 headlines and descriptions from \url{https://www.huffpost.com}, spanning 2012 to 2022~\cite{misra2021sculpting}. The "\textit{News Articles Dataset}" comprises articles from 2015 to 2017, scraped from \url{https://www.thenews.com.pk/}, focusing on business and sports~\cite{kaggle_news_articles}.  Larger corpora such as RealNews, constructed from CommonCrawl dumps, offer 120GB of deduplicated news articles from 2016 to 2019, targeting large-scale training and evaluation of LLMs~\cite{NEURIPS2019_3e9f0fc9}.  Additionally, the "\textit{20 Newsgroups}" dataset\footnote{\url{https://huggingface.co/datasets/google-research-datasets/newsgroup}}, widely used in NLP, contains approximately 20,000 documents grouped into 20 categories, though it is neither based on newspaper content nor does it specifically look at developments over time~\cite{Lang95}.

Regarding news datasets specifically for the German language, the "\textit{One Million Posts Corpus}" is one of the most prominent examples. It is derived from online discussions on the Austrian newspaper DER STANDARD’s website. It contains user posts from 2015-06-01 to 2016-05-31, with 11,773 labelled and 1,000,000 unlabelled entries, providing valuable insights into user-generated content~\cite{Schabus2017}. This is, however, notably different from our resource as we do not focus on user-generated content, but on the articles themselves. In addition to this, several linguistic newspaper corpora exist, offering access to German news data. These include DWDS~\cite{dwds}, the TüPP-D/Z corpus~\cite{TüPP-D/Z}, the Mannheim German Reference Corpus (DeReKo)~\cite{kupietz2009mannheim}, Leipzig Wortschatz~\cite{WortschatzLeipzig}, TIGER~\cite{tiger}, and others~\cite{instance1290, NoldaBarbaresiGeyken+2021+317+322}. However, these resources are often limited to keyword searches and return only sentence-level results, with most of them being unavailable for public use.

These datasets highlight the diversity of resources available for English and German news research but also reveal limitations, such as restricted access and narrow use cases. This underscores the need for openly available, comprehensive datasets like \texttt{taz2024full} to support language-specific studies and address gaps in research on German news media.

The detection of bias and discrimination in NLP has received significant attention in recent years. \citet{blodgett-etal-2020-language}, \citet{sun-etal-2019-mitigating}, and \citet{shah-etal-2020-predictive} provide broad overviews of existing approaches, highlighting the diverse methodologies employed to identify, mitigate, and evaluate biases in textual data. These works discuss various strategies, from detecting stereotypical associations in embeddings to evaluating fairness in predictive systems. Their surveys cover theoretical frameworks and practical implementations, offering valuable insights into the state of the field.

Despite the breadth of these reviews, none of the techniques discussed are directly comparable to the approach proposed by \citet{urchs-etal-2024-detecting}, which we adopt for evaluating our dataset. Unlike many conventional bias detection methods, which focus on linguistic patterns, embeddings, or statistical measures, the method of \citet{urchs-etal-2024-detecting} combines information extraction and linguistic discourse analysis to identify markers of bias and discrimination at the actor level. This actor-focused approach enables a more nuanced examination of how individuals and groups are represented in text, making it particularly suited for analysing bias in news media.

\section{Dataset}
We introduce the \texttt{taz2024full} newspaper corpus, a German newspaper corpus containing 1,834,370 publicly available articles published between 1980 and 2024 in the German newspaper "taz". This extensive dataset provides a valuable resource for linguistic, cultural, and societal research. It enables analyses across more than four decades of journalistic content. To the best of our knowledge, this is a unique collection of articles from a single news source over such an extended period, offering insights beyond specific use cases and opening avenues for long-term, diachronic studies (cf. \autoref{exp}).

The dataset covers a wide range of historical contexts, including events of global significance, such as 9/11, the financial crisis in Europe, and the COVID-19 pandemic, all of which have shaped discourse on an international scale. Additionally, it encompasses events with particular relevance to Germany, such as the reunion, the 2015 migrant crisis, and several political changes, providing researchers with a lens to explore national and regional impacts on public discourse. The corpus's temporal span allows for the analysis of how language, societal attitudes, and journalistic practices have evolved in response to these events.

For our corpus, we exclusively used "taz" as a data source because no other German newspaper granted us free access to their archives. Licensing fees would have been required, and even then, access to the full dataset would not have been guaranteed. Furthermore, publishing the corpus would have been prohibited due to legal restrictions on data redistribution.

We have chosen not to provide a predefined train-test split for the dataset. Data splitting strategies may vary significantly depending on research objectives and use cases. For instance, studies focusing on the impact of specific historical events may require custom temporal splits, while others analysing long-term trends might need broader, cross-temporal divisions. Allowing users to define their splits ensures maximum flexibility and adaptability for diverse research applications.

By offering a corpus that captures the breadth and depth of "taz" reporting across decades, we aim to provide a foundation for a wide range of studies, from exploring shifts in public discourse to examining linguistic phenomena and identifying patterns of bias and discrimination. This adaptability makes the \texttt{taz2024full} corpus a critical resource for researchers in NLP, computational social science (CSS), and related fields.

To our knowledge, this corpus is the largest German newspaper corpus available. Other publicly accessible German newspaper corpora do not provide full access to their data; instead, they typically allow only keyword-based searches or sentence-level queries through linguistic databases, making large-scale analysis impossible.

\subsection{Data Source}
The "taz" (\textit{die Tageszeitung}, \url{https://taz.de/}) is a German daily-occurring, left-leaning newspaper based in Berlin. First published on September 22, 1978, it transitioned to a daily publication schedule on April 17, 1979~\cite{taz_interview}. Known for its progressive editorial stance, "taz" has built a reputation as a prominent voice in the German media landscape, with 13,800 subscribers and an additional 39,000 paying for digital content~\cite{taz_numbers}. The newspaper plans to cease daily printing on October 17, 2025, transitioning to an online-only format while retaining a weekly Saturday print issue. This transition marks a shift reflective of broader trends in the news industry.

The "taz" newspaper is known for its diverse and comprehensive reporting, covering a wide array of topics. Its editorial structure is organised into various sections, including breaking news, politics, society, culture, and sports. Additionally, "taz" has dedicated segments such as "Öko," which covers economics, ecology, labour, consumption, transport, science, and the network economy. There are also regional sections focusing on Berlin and northern Germany, as well as "Wahrheit," which features unconventional content like satire, commentary, and creative formats. Unfortunately, the metadata we crawled did not include explicit labels indicating the section in which an article was published. As a result, this information could not be incorporated into the dataset. Nonetheless, the corpus reflects the full spectrum of "taz" journalism, encompassing a rich variety of topics and perspectives.

\subsection{Data Collection}
The \texttt{taz2024full} corpus was created by crawling publicly available content from the "taz" website (\url{https://taz.de/}) between August 2024 and November 2024. Permission has been granted to use and release this dataset for academic research, though its use for commercial purposes is strictly prohibited. The dataset only contains articles with more than three tokens (measured with SoMaJo tokeniser~\cite{Proisl_Uhrig_EmpiriST:2016}), thus excluding articles that only contain text fragments. The \texttt{taz2024full} corpus is publicly available on Zenodo\footnote{https://doi.org/10.5281/zenodo.15480855} for non-commercial purposes.

The articles are stored in JSON format. \autoref{fig:fig1} provides an example of the full JSON structure, illustrating how the metadata and article components are organised in the dataset. The metadata was extracted directly from the HTML of the crawled articles, and no modifications were made to the entries. The JSON format contains the following fields:

\begin{itemize} 
    \item \textbf{\texttt{"published\_on"}}: The publishing date, stored as a string in the format \texttt{YYYY-MM-DDThh:mm:ss+01:00}. 
    
    \item \textbf{\texttt{"contains\_actors"}}: A boolean indicating whether person entities were detected in the article. 
    
    \item \textbf{\texttt{"crawled\_on"}}: The crawling date, saved as a string in the format \texttt{YYYY-MM-DD hh:mm}. 
    
    \item \textbf{\texttt{"language"}}: The language of the article, always set to \texttt{"de"}. 
    
    \item \textbf{\texttt{"type"}}: Always set to \texttt{"article"}. 
    
    \item \textbf{\texttt{"author"}}: The person who wrote the article. 
    
    \item \textbf{\texttt{"keywords"}}: Keywords relevant to the article could be used for topic recognition. 
    
    \item \textbf{\texttt{"token\_count"}}: The number of tokens in the article.
\end{itemize}

The article-data consists of three parts, though not all are always available. However, every entry contains at least one \texttt{"text"} component:

\begin{itemize} 
    \item \textbf{\texttt{"title"}}: The headline of the article.
    
    \item \textbf{\texttt{"teaser"}}: A short description of the content or a short introduction.
    
    \item \textbf{\texttt{"text"}}: The main body of the article. 
\end{itemize}

\begin{figure}[h]
    \centering
    \begin{lstlisting}
{
    ID: {
        "meta_data": {
            "published_on": string,
            "contains_actors": boolean,
            "crawled_on": string,
            "language": "de",
            "type": "article",
            "author": string,
            "keywords": string,
            "token_count": int
        },
        "text": {
            "title": string,
            "teaser": string,
            "text": string
        }
    }
}
\end{lstlisting}
    \caption{Structure of the elements in the corpus, including all available metadata collected alongside the raw texts.}
    \label{fig:fig1}
\end{figure}

\subsection{Dataset Statistics}
\begin{figure*}[ht]
    \centering    
        \includegraphics[width=\textwidth]{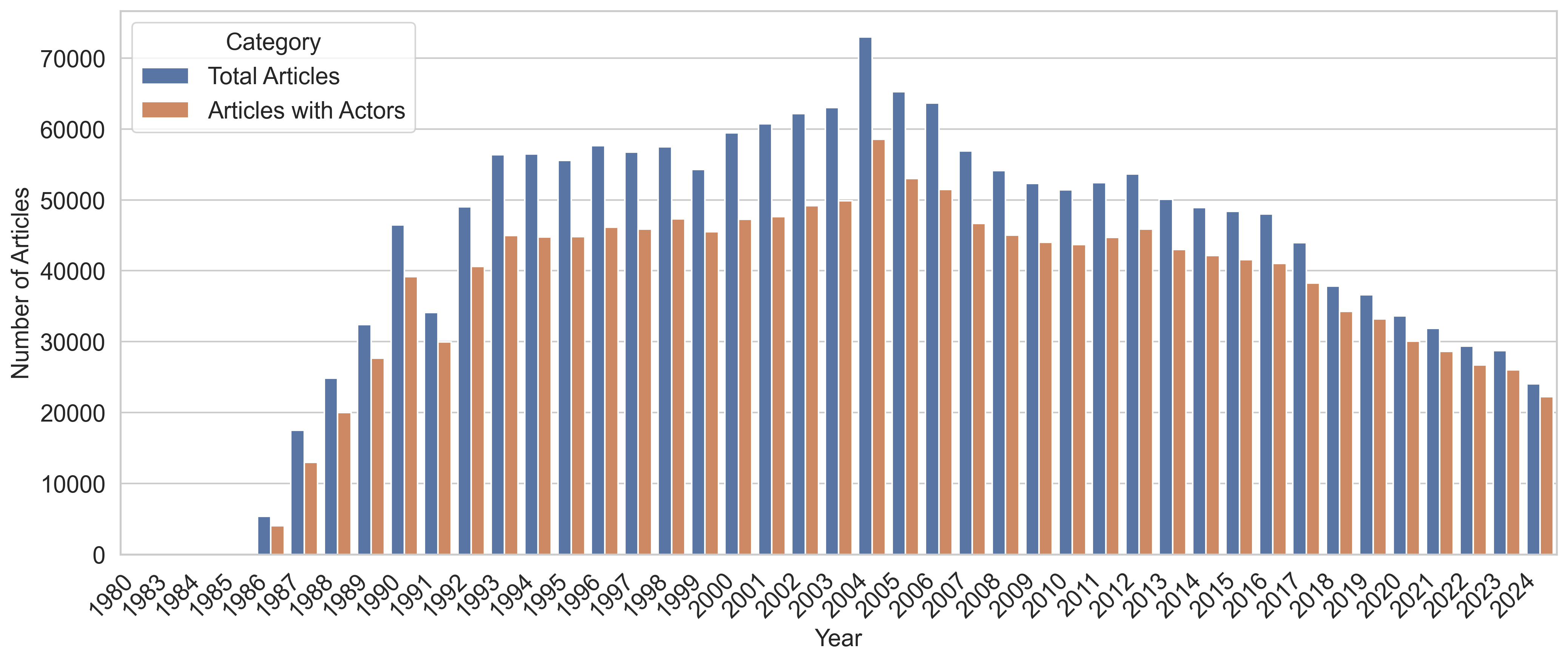}
        \caption{Number of newspaper articles and articles containing actors per year from 1980 to 2024. Years before 1986 include four articles or fewer.}
        \label{fig:article_year}
\end{figure*}

The \texttt{taz2024full} corpus consists of 1,834,026 newspaper articles published in the German newspaper taz between 1980 and 2024 (cf. \autoref{fig:article_year}). From 1980 onwards, there was a steady increase in the number of published articles, reaching a peak of 73,002 in 2004. An unexpected dip occurred in 1991, though the reason for this anomaly is unclear. After 1993, publication numbers stabilised again before declining after the peak in 2004. This decline may be linked to changes in the publishing strategy, potentially involving an increase in paid content from 2007 onwards. Only publicly available content is used for this corpus, thus leading to declining article numbers from 2007 onwards. 

The corpus contains 6,944,197 unique tokens, as determined using the SoMaJo tokeniser~\cite{Proisl_Uhrig_EmpiriST:2016}. Additionally, 83\% of all articles mention specific individuals, allowing for an analysis of gender bias and discrimination in how different actors are represented in the article. \autoref{tab:statistics} provides an overview of token, sentence, and article lengths. The maximum token and sentence lengths suggest that the tokeniser did not always ideally segment the articles. However, an average of four characters per token and 17 tokens per sentence appears reasonable.

\begin{table}
    \centering
    \begin{tabular}{p{1.8cm}|c|c|c|c}
     & min & max & mean & median \\ \hline
     token length & 1 & 2,238 & 5.15 & 4 \\ \hline
     sentence length & 1 & 18,872 & 20.07 & 17 \\ \hline 
     article length (token) & 3 & 26,855 & 396.89 & 276 \\ \hline
     article length (sentences) & 1 & 1,027 & 19.77 & 13 \\
    \end{tabular}
    \caption{Statistics for the \texttt{taz2024full} corpus.}
    \label{tab:statistics}
\end{table}

\subsection{Dataset Applications}
The \texttt{taz2024full} dataset offers numerous applications for researchers in NLP, CSS, and related fields. Its extensive temporal span supports investigations into the evolution of the German language, shifts in public discourse, and changes in thematic emphasis. For instance, it is possible to explore how language has adapted over decades, whether specific topics follow seasonal patterns, or how societal events have influenced reporting.

The dataset is particularly suited for examining patterns of bias and discrimination, mainly because it focuses on actors in texts. By leveraging this resource, insights can be obtained into representation trends and linguistic framing over time.

While the dataset is limited to a single source—the German newspaper taz, known for its progressive editorial stance—this consistency also ensures comparability across time. However, findings drawn from the dataset cannot be assumed to generalise across the broader German media landscape. We, therefore, recommend interpreting results in light of this context or using the dataset as a foundation for comparative analyses with other corpora.

By providing this comprehensive dataset, we aim to support a wide range of research efforts, from linguistic studies to bias detection, thereby contributing to a deeper understanding of German media discourse.

\section{Experiments}
\label{exp}

To showcase the utility of the \texttt{taz2024full} corpus, we conduct an illustrative analysis of gender representation in German newspaper articles over the past four decades. This analysis is not intended as an exhaustive study of bias, but rather as a demonstration of how the dataset can support research into long-term patterns of representation and framing.

We build upon the pipeline introduced by \citet{urchs-etal-2024-detecting}, which detects bias and discrimination at the actor level. While their work focused on English-language news, we adapt and scale an extended version of their pipeline to our German corpus, applying it to examine discrimination markers in journalistic reporting from 1980 to 2024.

The full code is published on GitHub\footnote{\url{https://github.com/Ognatai/corpus_pipeline}}. See also appendix \ref{reports} for a sample report of 2023.

\subsection{Data}
We analyse the entire corpus for discrimination markers to demonstrate its potential for bias and discrimination research.

The dataset encompasses all topics covered by "taz" without filtering, ensuring a comprehensive representation of the newspaper's reporting across politics, culture, economy, society, and more. 

The chosen time-frame enables a meaningful comparison, capturing how taz’s reporting on actors has evolved over 44 years. This period includes significant societal, cultural, and journalistic developments, which are likely to influence how gender bias and discrimination manifest in the texts. 

\subsection{Method}
\citet{urchs-etal-2024-detecting} automate linguistic discourse analysis by applying information extraction techniques to analyse English newspaper articles for discriminatory content. This approach focuses on two key aspects of linguistic discourse analysis: nomination, which examines how actors in a text are named, and predication, which analyses how they are described.

Their pipeline is designed to identify actors within the text by leveraging named entity recognition (NER) to detect person entities. The authors utilise spaCy\footnote{\url{https://spacy.io/}} to extract all named entities from the text, retaining only those classified as persons. Additionally, generic terms referring to individuals, such as "mother," "father," "woman," and "man", are extracted. Entities that share the same name or name components (e. g. if only a first name is present or only a surname) are grouped together under a single actor. To further refine the analysis, coreferee\footnote{\url{https://spacy.io/universe/project/coreferee}} is employed for coreference resolution, linking pronouns to the corresponding actors. This approach enables text analysis based on distinct actors rather than relying solely on pronoun frequency.

The pipeline subsequently assumes the gender of the actors based on the extracted pronouns, categorising them as \texttt{woman}, \texttt{man}, \texttt{non-binary}, or \texttt{undefined}. If more than 70\% of the pronouns associated with an actor are either feminine or masculine, the actor is categorised as \texttt{woman} or \texttt{man}, respectively. Actors are labelled \texttt{undefined} if no pronouns are present or the thresholds for other categories are not met.

\citet{urchs-etal-2024-detecting} extract sentences mentioning the actors to analyse their predication. Therefore each sentence that either contains the name of the actor or a pronoun linked to the actor is extracted. This data is used to identify the following markers of discrimination across the gender categories:
\begin{compactitem}
    \item Number of actors in text per gender category
    \item Count of mentions per gender category / individual
    \item Sentiment towards gender category / individual\footnote{The spaCy pipeline component spacytextblob (\url{https://spacy.io/universe/project/spacy-textblob}) is used for sentiment analysis. In our German version we use the pre-traind model "oliverguhr/german-sentiment-bert", which is trained on German newspaper articles, from huggingface (\url{https://huggingface.co/transformers/v3.0.2/main_classes/pipelines.html})}
    \item Count of feminine-coded words and masculine-coded words in the predication of each gender category/individual\footnote{We utilise the lists of feminine- and masculine-coded words provided on the Gender Decoder website \url{https://www.msl.mgt.tum.de/rm/third-party-funded-projects/projekt-fuehrmint/gender-decoder/wortlisten/}, which are derived from the work of Gaucher et al.~\cite{gaucher2011evidence}. Their research demonstrates that women are more frequently associated with communal traits, while men are linked to agency-related terms.}
    \item Abusive words in predications (not used in this work)
\end{compactitem}
These markers are then compiled into a discrimination report and visually represented for each text. 

In this work, we adapt the existing pipeline to analyse German newspaper articles, expanding its scope from processing individual articles to handling thousands at a time. Our adaptations include the following:
\begin{itemize}
    \item \textbf{Gender Assumption:} Instead of assuming the gender of actors, we determine the primary pronoun used to refer to them. Additionally, we scanned the corpus for German neo-pronouns\footnote{We used the neo-pronouns listed at \url{https://gleichstellung.tu-dortmund.de/projekte/klargestellt/neo-pronomen}} and found that only five texts contained such pronouns. Consequently, our analysis focuses on the pronouns she/her and he/him. For the results section, we categorise actors primarily referred to with she/her pronouns as women and those with he/him pronouns as men.

    \item \textbf{Pronoun Driven Analysis:} We include only actors for whom co-reference with pronouns could be established. All other actors are not part of the analysis.

    \item \textbf{Generic Masculine/German Gender-Neutral Language:} We introduce a marker to determine whether an article employs the generic masculine and one for the German gender-neutral language.

    \item \textbf{Pairwise Mutual Information (PMI):}  Additionally, we identify the top 10 adjectives with the highest PMI in each actor's predication. PMI (cf. equation \ref{pmi}) measures the probability of two words $x$ and $y$ co-occurring by chance or meaningfully. A higher PMI indicates a more meaningful relation between these words~\cite{jurafsky2000speech}. In our pipeline, we calculate the PMI for each actor and each adjective in their predication, excluding stop-words. Therefore, we can identify the most influential adjectives for each actor.

    \begin{equation}
        \textsf{PMI}(x, y) = log_2 \frac{P(x, y)}{P(x)P(y)}
        \label{pmi}
    \end{equation}

    \item\textbf{Aggregated Report:} We introduce a human-readable report to facilitate easy corpus analysis.
\end{itemize}

We deliberately chose not to use large language models (LLMs) for the analysis, despite known challenges with pronoun detection and co-referencing. While LLMs could potentially enhance performance, we opted against them due to concerns about the biases they may introduce and the lack of transparency in their outputs. Instead, we relied on well-established, interpretable methods, acknowledging their limitations. This decision prioritises ethical considerations and ensures greater methodological control.

We conduct a yearly analysis of the whole corpus, examining changes and trends over the 44-year span of our corpus.
 
\subsection{Results}
\begin{figure}[ht]
    \centering 
    \includegraphics[width=0.5\textwidth]{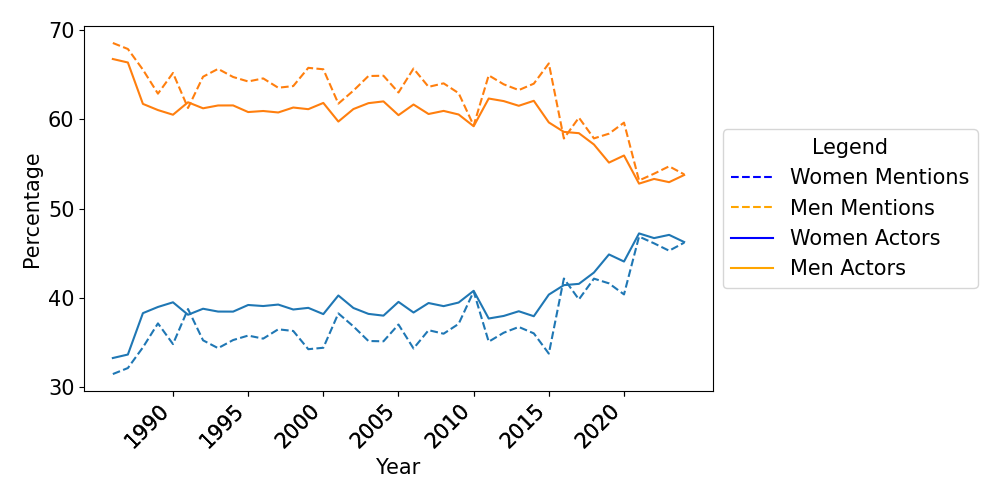}
    \caption{Comparison of the number of actors per article based on detected genders and the frequency of their mentions within each article.}
    \label{fig:actors_mentions} 
\end{figure}

The analysis of the corpus over the years provides valuable insights into how taz has written about women and men across the decades. \autoref{fig:actors_mentions} illustrates the proportion of actors whose gender could be co-referenced with pronouns, distinguishing between women and men (Woman Actors/Man Actors). Additionally, the figure presents the frequency with which these actors are mentioned within texts (Woman Mentions/Man Mentions)\footnote{We differentiate between the number of actors in a text and the number of mentions to account for cases where, for instance, a single woman actor is referenced multiple times, whereas multiple male actors might be mentioned only once. Without this distinction, a text featuring one woman mentioned ten times and another featuring ten men mentioned once each would appear equivalent in terms of gender representation, despite their differing narrative emphases.}. Before 1990, the data is too sparse to allow definitive interpretations. However, from the 1990s onward, a clear trend emerges: "taz" reported significantly more on men than on women. This imbalance is present across all decades, with man actors not only more frequently included in articles but also mentioned more often. While a shift towards greater inclusion of women actors becomes apparent from the 2010s onwards, this pattern of men actors being both more often the subject of reporting and more frequently referenced persists. Even in recent years, where gender representation appears almost balanced in terms of actor inclusion, men continue to receive more textual space, indicating a continued dominance in media visibility.

\begin{figure}[ht]
    \centering 
    \includegraphics[width=0.45\textwidth]{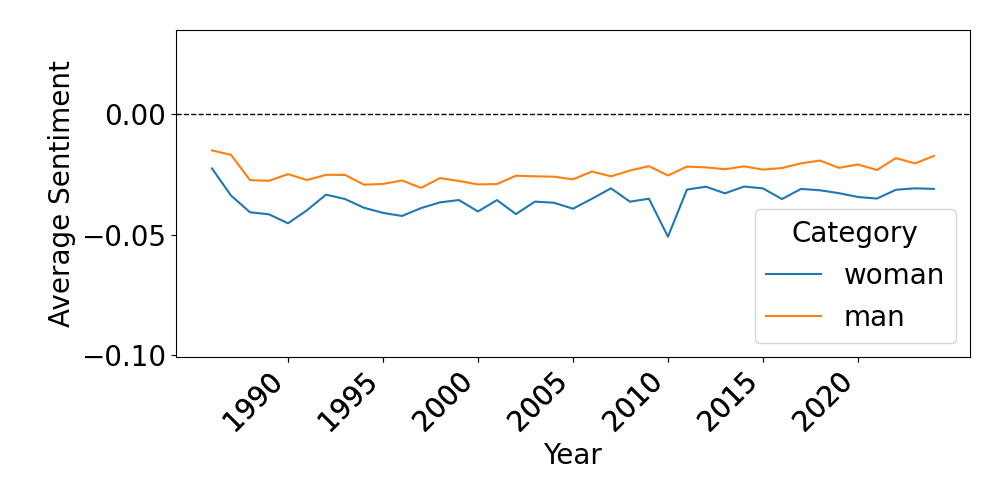}
    \caption{Sentiment towards the detected genders through the years.}
    \label{fig:sentiment} 
\end{figure}

Beyond the quantity of mentions, the sentiment towards men and women in "taz" articles is also revealing. \autoref{fig:sentiment} illustrates the sentiment associated with women and men actors over time. Sentiment values range from -1 (highly negative) to +1 (highly positive), with 0 representing neutrality. The data shows that "taz" articles generally lean towards a neutral but slightly negative sentiment. More strikingly, across the entire 44-year period, sentiment towards women actors is consistently slightly more negative than sentiment towards men actors. While the differences are not extreme, this persistent pattern suggests that women in "taz" articles are, on average, portrayed in a slightly more negative light than their male counterparts.

Looking at language use more closely, an analysis of adjectives with the highest PMI association with women and men actors reveals that these descriptors remain relatively stable over time and do not exhibit strong gender differentiation. Additionally, a targeted analysis of female-coded and male-coded words shows that these were used only rarely. This suggests that "taz" makes little use of explicitly gender-coded language. However, our analysis found no evidence that "taz" systematically uses German gender-neutral language. While a few instances of gender-neutral forms were manually observed, these were rare exceptions rather than a common practice. Despite ongoing discussions about inclusive language in German, "taz" does not appear to have adopted gender-neutral writing conventions in its standard editorial style.

\section{Conclusion and Future Work}

In this work, we introduced \texttt{taz2024full}, a comprehensive German newspaper corpus spanning over four decades. This dataset represents a significant resource for linguistic, cultural, and societal research, particularly in the areas of gender bias and discrimination in media. By leveraging a structured approach to analysing gender representation through actor mentions and predications, we provided insights into how "taz" has reported on women and men over time. Our findings highlight a persistent imbalance in gender representation, with men not only appearing more frequently as actors but also receiving greater textual space.

Furthermore, we demonstrated the adaptability of an existing bias detection pipeline, originally designed for English texts, to large-scale German-language data. Our extensions included modifications tailored to the German linguistic landscape, such as pronoun-based gender identification and an analysis of the generic masculine. These enhancements offer a more refined approach to studying gender-related language use in German news media.

Looking ahead, several avenues for future research emerge. One priority is updating the corpus with new data beyond 2024 to enable ongoing diachronic analysis. Additionally, incorporating topic modelling could provide deeper insights into the contextual framing of gender representation. Since the current implementation would model topics at the sentence level, a necessary improvement would involve incorporating broader textual context to enhance topic coherence around actor predications.

Another promising direction is argument mining, which could refine our understanding of the implicit and explicit biases embedded in journalistic discourse. By identifying argument structures and rhetorical strategies, we could further uncover how gender bias manifests in media narratives.

Ultimately, we hope that \texttt{taz2024full} serves as a valuable resource for researchers in NLP, computational social science, and related fields, facilitating future studies on bias, representation, and media discourse in the German language.

The \texttt{taz2024full} corpus and its language-agnostic pipeline provide a foundation for analysing bias and discrimination in German news media. Future work will focus on expanding the corpus with newer data to ensure it remains relevant and reflective of contemporary discourse. Enhancements to the pipeline include training sentiment analysis models tailored to German newspapers, improving pronoun coreference resolution for German texts, and incorporating argument mining for discrimination detection. Additionally, we aim to extend the pipeline to support multimodal analysis by integrating text with non-textual data such as images and videos, enabling a comprehensive understanding of media content.

\section*{Use of AI}
The authors are not native English speakers; therefore, ChatGPT and Grammarly were used to assist with writing English in this work.

\section*{Limitations}
The \texttt{taz2024full} corpus has several limitations to consider when interpreting research results. Firstly, the dataset reflects the views of a Berlin-based, left-leaning publication and does not represent the entire spectrum of German discourse. This inherent bias limits its applicability for studying nationwide or ideologically diverse perspectives.

Additionally, bias and discrimination detection within corpora is inherently subjective, as no universally accepted gold standard exists. Different users may have varying values and interpretations of bias or discrimination, complicating evaluating such tasks. Therefore, we are not able to decide if a corpus is discriminatory or biased, as we do not know the use cases for each corpus. Additionally, our understanding of discrimination or bias might differ from the users understanding. Thus, we limit the output to a discrimination report, allowing each user to determine whether any adjustments to the corpus are necessary or to compare different corpora based on the calculated metrics.

Further challenges arise when applying the language-agnostic, flexible pipeline to detect gender discrimination and bias in texts. Co-reference resolution, particularly on German data, remains problematic due to the lower accuracy of current models for this language. This can affect the precision of gender-related analyses. Furthermore, the lack of directly comparable works or benchmarks complicates evaluating the pipeline's performance.

\section*{Ethical Considerations}
The \texttt{taz2024full} corpus is intended exclusively for academic research purposes, and exploiting it for commercial use would harm the publisher, taz Verlags und Vertriebs GmbH. To prevent such misuse, we strongly emphasise the importance of adhering to the dataset's intended purpose: fostering academic exploration and understanding.

Defining discrimination can be addressed abstractly, but implementing the concrete pipeline requires concrete values and definitions. Since we do not know the pipeline users' specific use cases and value systems, we opted for a flexible analysis of discrimination markers, highlighting potentially problematic content, thus refraining from definite judgments. We intentionally leave the final interpretation to human users. This allows them to apply their understanding of what constitutes discrimination in their specific use cases.

We acknowledge that our pipeline could be misused to curate datasets with specific biases or intentionally exclude particular genders. We strongly discourage using this system to manipulate datasets in ways that reinforce or amplify discrimination and biases. Instead, we aim to promote fairness and inclusivity by providing insights that help users curate discrimination and bias-free data sets.

By maintaining human oversight in the evaluation process, we aim to balance automated analysis with ethical responsibility, ensuring that the system supports diverse needs while promoting fairness in language technologies.

\section*{Acknowledgements}
This work was written by an author team working in different projects. Stefanie Urchs’ project “Prof:inSicht” is promoted with funds from the Federal Ministry of Education and Research under the reference number 01FP21054. Matthias Aßenmacher is funded with funds from the Deutsche Forschungsgemeinschaft (DFG, German Research Foundation) as part of BERD@NFDI - grant number 460037581. Responsibility for the contents of this publication lies with the authors.

\bibliography{corpora_discrimination_detection}

\clearpage
\newpage

\appendix
\section{Corpus Report 2023}
\label{reports}

\noindent\begin{minipage}{\textwidth}
    \centering
    \lstset{
        numbers=none,
        language=,
        basicstyle=\scriptsize\ttfamily,
        frame=none
    }
    \begin{lstlisting}
Aggregated Report for 2023
============================================================

Total Texts: 26357
Total Actors: 17161
Pronoun Distribution: {'he_him': 9088, 'she_her': 8073}
Total Mentions: 109634
Mentions by Pronoun: {'he_him': 60008, 'she_her': 49626}

Mean Metrics:
  total_actors: 1.67
  total_mentions: 10.66
  total_feminine_coded_words: 0.27
  total_masculine_coded_words: 0.11
  contains_majority_gender_neutral: 0.00
  generic_masculine: 0.31
  pronoun_distribution_she_her: 0.79
  pronoun_distribution_he_him: 0.88
  mentions_pronoun_distribution_she_her: 4.83
  mentions_pronoun_distribution_he_him: 5.84
  feminine_coded_words_pronoun_distribution_she_her: 0.13
  feminine_coded_words_pronoun_distribution_he_him: 0.14
  masculine_coded_words_pronoun_distribution_she_her: 0.05
  masculine_coded_words_pronoun_distribution_he_him: 0.06
  average_sentiment_all: -0.03
  sentiment_by_pronoun_she_her: -0.03
  sentiment_by_pronoun_he_him: -0.02

Median Metrics:
  total_actors: 1.00
  total_mentions: 5.00
  total_feminine_coded_words: 0.00
  total_masculine_coded_words: 0.00
  contains_majority_gender_neutral: 0.00
  generic_masculine: 0.00
  pronoun_distribution_she_her: 1.00
  pronoun_distribution_he_him: 1.00
  mentions_pronoun_distribution_she_her: 2.00
  mentions_pronoun_distribution_he_him: 2.00
  feminine_coded_words_pronoun_distribution_she_her: 0.00
  feminine_coded_words_pronoun_distribution_he_him: 0.00
  masculine_coded_words_pronoun_distribution_she_her: 0.00
  masculine_coded_words_pronoun_distribution_he_him: 0.00
  average_sentiment_all: 0.00
  sentiment_by_pronoun_she_her: 0.00
  sentiment_by_pronoun_he_him: 0.00
 
Top PMI Adjectives Table:
All                  she/her              he/him              
--------------------------------------------------------------------------------
letzten              junge                letzten             
deutschen            letzten              russischen          
junge                deutschen            deutschen           
russischen           jungen               politische          
berliner             deutsche             berliner            
deutsche             berliner             politischen         
politische           russischen           ukrainischen        
politischen          nächsten             ukrainische         
jungen               alten                russische           
nächsten             alte                 deutsche            
    \end{lstlisting}
    \captionof{lstlisting}{Full discrimination report of 2023.}
    \label{fig:report_2023}
\end{minipage}

\end{document}